\documentclass[journal]{IEEEtran}
\ifCLASSINFOpdf
  \usepackage[pdftex]{graphicx}
  \graphicspath{{pdf/}{jpeg/}}
  \DeclareGraphicsExtensions{.pdf,.jpg,.jpeg,.png}
\else
  \usepackage[dvips]{graphicx}
  \graphicspath{{eps/}}
  \DeclareGraphicsExtensions{.eps}
\fi
\usepackage{amsthm}
\usepackage{amsmath}
\usepackage{amssymb}
\usepackage{bbding}
\usepackage{multirow}
\usepackage{booktabs}
\usepackage{threeparttable}
\usepackage{subfig}
\usepackage[dvipsnames, svgnames, x11names]{xcolor}

\begin{document}
\title{Weakly Supervised Temporal Adjacent Network for Language Grounding}

\author{Yuechen~Wang,
        Jiajun~Deng,
        Wengang~Zhou,
        and~Houqiang~Li,~\IEEEmembership{Fellow,~IEEE}
\thanks{Yuechen~Wang, Jiajun~Deng, Wengang~Zhou and Houqiang Li are with the Department of Electrical Engineering and Information Science, University of Science and Technology of China, Hefei, 230027 China (e-mail: wyc9725@mail.ustc.edu.cn, dengjj@mail.ustc.edu.cn, zhwg@ustc.edu.cn, lihq@ustc.edu.cn).}
\thanks{Corresponding authors: Wengang Zhou and Houqiang~Li.}
\thanks{This work is supported by the GPU cluster built by MCC Lab of Information Science and Technology Institution, USTC.}}

\maketitle

\begin{abstract}
Temporal language grounding (TLG) is a fundamental and challenging problem for vision and language understanding. Existing methods mainly focus on fully supervised setting with temporal boundary labels for training, which, however, suffers expensive cost of annotation.
In this work, we are dedicated to weakly supervised TLG, where multiple  description sentences are given to an untrimmed video without temporal boundary labels. 
In this task, it is critical to learn a strong cross-modal semantic alignment between sentence semantics and visual content.
To this end, we introduce a novel weakly supervised temporal adjacent network (WSTAN) for temporal language grounding. 
Specifically, WSTAN learns cross-modal semantic alignment by exploiting temporal adjacent network in a multiple instance learning (MIL) paradigm, with a whole description paragraph as input. 
Moreover, we integrate a complementary branch into the framework, which explicitly refines the predictions with pseudo supervision from the MIL stage.
An additional self-discriminating loss is devised on both the MIL branch and the complementary branch, aiming to enhance semantic discrimination by self-supervising.
Extensive experiments are conducted on three widely used benchmark datasets, \emph{i.e.}, ActivityNet-Captions, Charades-STA, and DiDeMo, and the results demonstrate the effectiveness of our approach.
\end{abstract}


%
\IEEEpeerreviewmaketitle

\section{Introduction}
\IEEEPARstart{T}{emporal} language grounding (also named as moment retrieval) aims to localize a temporal segment in a video corresponding to a language query.
Automatic temporal language grounding enables us to efficiently find the video moments of interest rather than going through the whole video, which is fundamental to various multi-modal tasks, \emph{e.g.}, visual question answering~\cite{VQA,8988148,MovieQA,8811730}, visual reasoning~\cite{CLEVRAD,COG}, video captioning~\cite{7984828, 9121763, 8031355} and storytelling~\cite{huangStorytelling,VideoStory}.
Since its first introduction~\cite{TALL,DiDeMo}, substantial efforts have been made on this problem~\cite{ijcai2018-143,Ge2019MACMA,aaai2019Yuan,MAN,2D-TAN,He2019ReadWA,cvprWangHW19}.

Early works approach this task mainly by supervised learning. In spite of tremendous achievements, the fully supervised methods need laborious manual annotation of temporal boundaries for every sentence query for training. Besides, the temporal labels annotated by different annotators are usually ambiguous and noisy because of subjectivity, damaging the learning of models. 
On the other hand, it is much easier to collect a large amount of video-level descriptions without temporal annotations, as this form of data is widely available on the Internet~(\emph{e.g.,} YouTube).
Based on such motivation, prior works~\cite{TGA,WSLLN,SCN} turn to the weakly supervised setting of temporal language grounding task, where several description sentences are given to an untrimmed video, while the temporal boundary labels are not provided.
To achieve temporal grounding, it is critical to learn cross-modal semantic alignment between sentence semantics and visual contents. However, existing weakly supervised methods learn the cross-modal semantic alignment by similarity measurement or textual semantic completion, which is not temporally discriminative enough.
In fact, due to the inherent difficulty to learn diverse semantic contents in videos with both high precision and recall, these weakly supervised methods are often degenerated to the retrieval of clips containing distinct semantic contents, failing to model the semantic context that is essential to fix temporal boundaries, \emph{e.g.}, small objects and postures.

\begin{figure}[tb]
	\centering
	\includegraphics[scale=0.425]{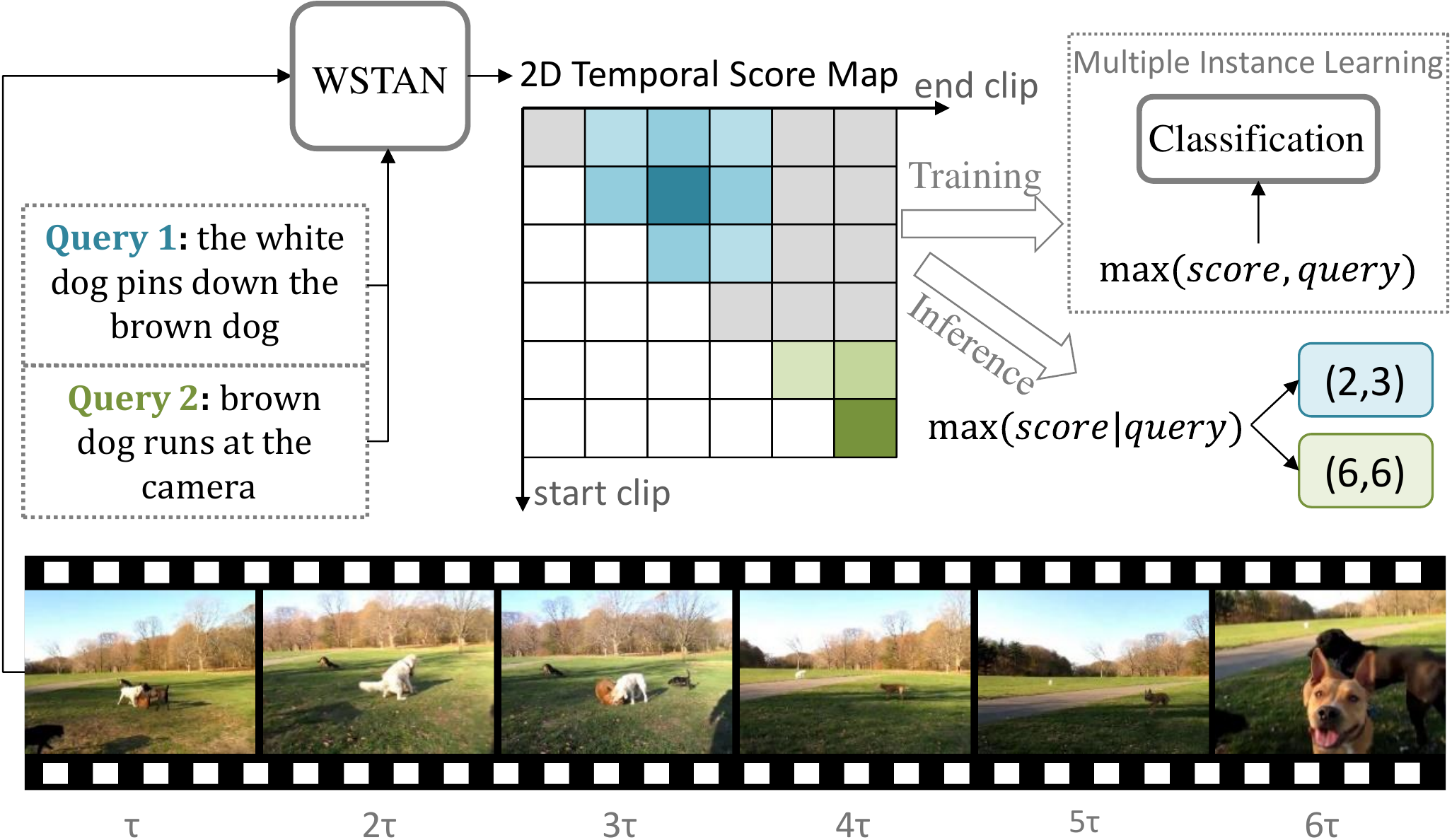}
	\caption{Illustration of the training and inference pipeline of WSTAN.
	With a video and a group of text descriptions as input, WSTAN produces 2D temporal scores maps, indicating the relevance of text queries and candidate video segments. 
	During training, the score maps are processed with max operation and a matching score is produced for cross-modal matching classification.
	For inference, the description sentences are considered separately and the moment on the score map which gives max response for a query is taken as grounding result.
	}
	\label{fig:fig1}
\end{figure}

In this paper, towards effective cross-modal alignment learning, we propose a novel framework, named weakly supervised temporal adjacent network (WSTAN).
The key idea of our approach is to treat the weakly supervised task as a multiple instance learning (MIL) problem to learn cross-modal alignment, as well as explicitly select and refine the prediction from MIL network. The explicit adjustment can help the model uncover more semantic concepts and be more temporally discriminative.
As shown in Fig.~\ref{fig:fig1}, we first construct a 2D feature map by aggregating features of evenly divided clips, and the $(i,j)^{th}$ element of the 2D map represents the video segment from the $i^{th}$ clip to the $j^{th}$ clip. The video segments on the 2D map are candidate proposals for subsequent network.
Then we concatenate all the description sentences for a video, and take the resulting text paragraph as a sample unit to perform cross-modal matching classification.
Specifically, we build the classifier by viewing candidate proposals as bag of instances in a multiple instance learning paradigm, and perform binary matching classification with video-level matching label.

To better model the relationships among candidate proposals, we exploit temporal adjacent network~\cite{2D-TAN} in the context of MIL with a whole paragraph as input.
Moreover, inspired by the success achieved by multi-branch methods on weakly supervised object detection problem~\cite{OICR}, we devise a complementary branch to compensate for the semantic sparsity in videos. The additional branch explicitly refines the predictions of base alignment network by taking the intermediate results as pseudo supervision. Therefore, it can bring up more semantically meaningful clips that are not distinct for matching classification.
We further leverage a self-discriminating loss on the 2D score map, which enables our model to be more temporally discriminative by self-supervising, emphasizing more semantic contents, especially activities.

In summary, the main contributions of our work are threefold as follows:
\begin{itemize}
	\item We propose a novel weakly supervised temporal adjacent network (WSTAN) for language grounding, which also leads to an elegant view to leverage the temporal structure information to formulate the weakly supervised temporal language grounding problem as an MIL problem.
	\item By integrating an additional complementary branch into the MIL framework, as well as leveraging self-discriminating loss, WSTAN learns cross-modal semantic alignment more precisely.
	\item Through comprehensive experiments, we demonstrate that our WSTAN outperforms several state-of-the-art methods on three widely adopted benchmarks, \emph{i.e.,},  ActivityNet-Captions~\cite{ActivityNet}, Charades-STA~\cite{TALL}, and DiDeMo~\cite{DiDeMo}.
\end{itemize} 

\section{Related Work}
In this section, we briefly review the related works for object detection, temporal action localization, and temporal language grounding.

\subsubsection{Object Detection} 
Object detection aims to localize and identify objects in an image, which is an important problem in computer vision, and has many applications. 
In supervised setting with object annotations~\cite{ImageNet,Pascal,MSCOCO}, great performance advance has been made over years~\cite{RCNN,Girshick2015FastR,FasterR,Liu2016SSDSS,YOLO,8125749,9156611,DETR}.
However, it is time-consuming to collect detailed annotations with bounding boxes, while image level object tags is easier to annotate. 
Based on such a motivation, many existing methods has explored the object detection problem under image level weak supervision.
Weakly Supervised Deep Detection Network (WSDDN)~\cite{WSDDN} presents an MIL framework, and performs region selection and classification simultaneously. Later, instance classifier refinement is integrated into the MIL framework~\cite{OICR}. 
Since then, multiple works were proposed to improve the pseudo supervise strategy of OICR~\cite{PCL,OIM}.
A combination of instance classifier refinement and bounding-box regression is further explored by adding an additional regression branch~\cite{Yang2019TowardsPE,WSOD2,Ren2020InstanceawareCA}.

Different from object detection, language grounding is a new task that tries to temporally localize a specific moment conditioned by a text sentence query. However, the methodology in object detection is inspiring to the advance of temporal language grounding methods.

\subsubsection{Temporal Action Localization}
Temporal action localization aims to detect a limited set of actions in untrimmed videos and localize the start and end frames of detected actions. Since the first introduction~\cite{Gaidon11actomsequence}, 
lots of efforts have been made in this area under both fully supervised and weakly supervised settings.
Most of the fully supervised works are designed in a multi-stage manner: first select temporal  proposals from candidate set and perform regression, then predict corresponding action labels from the pre-defined set~\cite{Zhao17,Lin17,Chao2018RethinkingTF}. Non-Maximum Suppression (NMS) is often applied as post-processing.
Segment-CNN~\cite{SCNN} generates multi-scale video segment and performs proposal selection, action classification and localization though a three-stage framework sequentially. 
To improve proposal precision for long video segments and decrease memory cost of sliding window, some efforts are devoted to strategy of generating candidate proposals~\cite{TAG,TURN-TAP}.

On the other hand, under the weakly supervised setting, only video-level action labels are available, which is much more challenging. UntrimmedNet~\cite{UntrimmedNet} presents a multiple instance learning (MIL) framework to perform action classification and attention-based video segment selection. In~\cite{Singh2017HideandSeekFA}, a hide-and-seek approach is applied in training, forcing the network to learn all the relevant frames. STPN~\cite{STPN} further enhances the performance by using both temporal class activations and class-agnostic attentions. Shou \emph{et al.}~\cite{AutoLoc} develop a novel framework to directly predict the temporal boundary of each action instance. In~\cite{Nguyen2019WeaklySupervisedAL}, both foreground and background frames are explicitly modeled, and a background-aware loss allows the network to learn richer representations for actions.

\subsubsection{Temporal Language Grounding} 
To expand action localization into more diverse applications, temporal language grounding is proposed~\cite{TALL,DiDeMo} as a new challenging task, requiring deep interactions between two modalities.
Previous methods have explored this task in a fully supervised setting~\cite{TALL,DiDeMo,ROLE2018,ABLR2018,Ge2019MACMA,Xu2019MultilevelLA,aaai19Chen,aaai2019Yuan,SIGIRZhang,MAN}. Most of them follow a two-stage paradigm: selecting candidate moments with sliding windows and subsequently matching the language query.
MAN~\cite{MAN} exploited the graph-structured moment relations with an iterative graph adjustment network for video representation learning. 
Recently, one-stage framework has been developed.
In 2D-TAN~\cite{2D-TAN}, a one-stage method called 2D Temporal Adjacent Network(2D-TAN) is proposed to model the context in candidate video moments. Recently, reinforcement learning  has been leveraged for temporal language grounding~\cite{He2019ReadWA,cvprWangHW19}. Through well-designed action space consisting of different ways to adjust the temporal boundaries, these methods avoid sliding over
the entire video, and achieve high detection speed. 

Despite the boom of fully supervised methods, it is very time-consuming and labor-intensive to acquire large number of temporal boundary annotations for supervision. And due to the annotation inconsistency among annotators, temporal labels may be confusing for models to learn. 
To alleviate the dependence of fine-grained annotation, weakly supervised setting is explored lately.
TGA~\cite{TGA} exploits Text-Guided Attention to map video and text features into a latent space to learn cross-modal similarity. WSLLN~\cite{WSLLN} learns segment-text matching and conducts segment selection simultaneously. Masked sentence complementary is also explored~\cite{SCN}. These methods learn cross-modal semantic alignment without any temporal instruction, struggling to recognize visual contents that are important for localization.
Differently, our WSTAN adopts pseudo labels to explicitly refine the predicted temporal boundaries, which allows the model to learn more concepts that are essential for localization.

\section{Our Approach}	

\begin{figure*}[htb]
	\centering
	\includegraphics[scale=0.485]{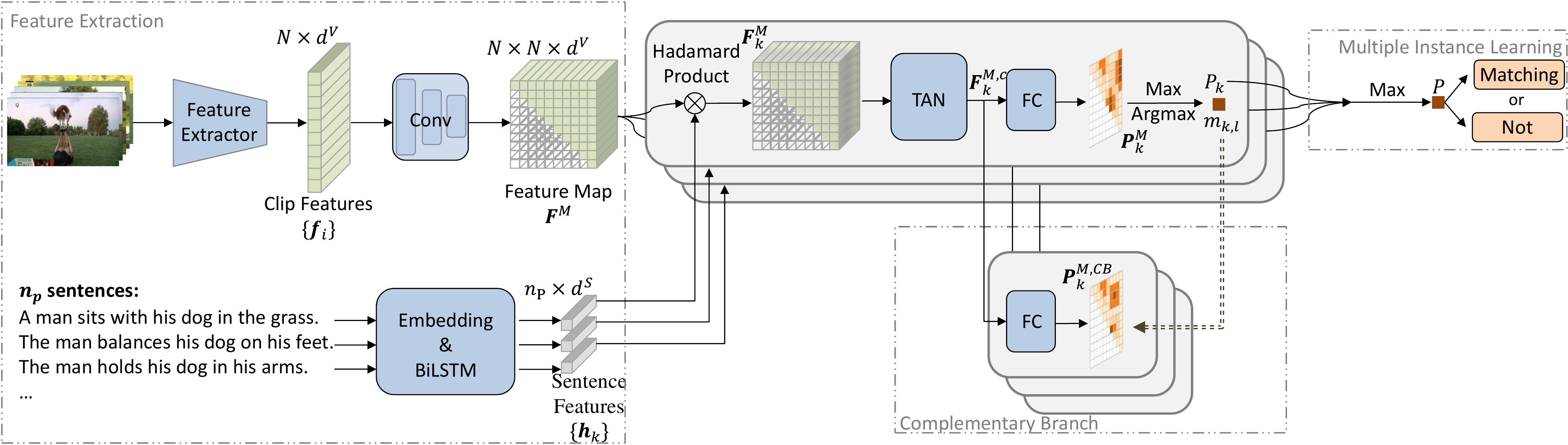}
	\caption{The architecture of WSTAN. 
		A 2D feature map is first extracted from the input video clips, and fuse with text features extracted by BiLSTM for each sentence separately. The fused features are processed through temporal adjacent network(TAN) to model relationships with adjacent moments, and then branch into two streams: cross-modal matching classifier, and complementary branch. Self-discriminating loss is adopted on predictions of both steams.}
	\label{fig:model_arch}
\end{figure*}

In this section, we first give a formal definition for temporal language grounding, and then describe the feature extraction procedure of both the visual and the linguistic modality. After that, we introduce our proposed WSTAN, consisting of cross-modal matching classifier, self-discriminating loss, and complementary branch. Last but not least, we discuss how to generate the temporal grounding results in the online inference stage. 

\subsection{Problem Formulation}
Given an untrimmed video and a text-sentence query, a temporal grounding model aims to localize the most relevant moment in the video, represented by its start and end timestamps. 
In this paper, we consider a weakly supervised setting, that is, for each video $\textbf{V}$, a set of text description sentences $\textbf{S}=\{S_k\}_{k=0}^{n_p-1}$ are provided for training, where $n_p$ is the number of sentences. Each sentence describes a specific moment in the video, yet the temporal boundaries are not provided for training.
In the inference stage, the weakly supervised trained model is required to predict the start and end time $\hat{\tau}=[\hat{s}, \hat{e}]$ of an input query $S$.

\subsection{Feature Extraction}
\subsubsection{Textual Feature}
For an untrimmed video, several description sentences are provided, each corresponding to a specific moment in the video.
We concatenate all sentences describing the input video into one description paragraph as positive textual input, and a description paragraph of another video is taken as the negative matching sample for the MIL procedure. And we randomly drop one sentence in the text paragraph. Therefore, the influence of simple descriptions for common events in videos is diluted and the feature extracting network is forced to learn semantics of every sentence, \emph{i.e.,} the recall of semantics in the textual input is improved. 

We extract textual feature for each sentence in paragraph $\{S_k\}_{k=0}^{n_p-1}$ separately, and obtain a series of features $\{\textbf{h}_k\}^{n_p-1}_{k=0}$. 
Specifically, for each word $w_j, j\in \{0, 1, \cdots, n_k\}$ in sentence $S_k$, we first embed it into $\textbf{w}_j \in \mathbb{R}^{d^S}$ using the Glove word2vec model~\cite{Glove}, where $d^S$ is the vector length. 
Then, we feed the word embeddings $\{\textbf{w}_j\}_{j=0}^{n_{k}-1}$ into a bidirectional LSTM network~\cite{LSTM}. The final hidden state of the BiLSTM $\textbf{h}_k \in \mathbb{R}^{d^S}$ is taken as the representation of sentence $S_k$ to interact with visual information. 

\subsubsection{Visual Feature}
Given an input video $\textbf{V}$, our goal is to construct candidate moments and extract their features for subsequent cross-modal fusion. 
To construct candidate moments, we build a 2D temporal map, where the two dimensions represent start and end time of the candidate moment, respectively. 
Specifically, we evenly divide the input video into $N$ clips. Then any video segment corresponding to a sentence can be viewed as a combination of several sequential clips, with acceptable loss. 
Therefore, a candidate moment can be identified by its start and end clips, and all possible candidates can be organized into a 2D temporal map $M$.
The element $m_{ij}$ in $M$ represents the candidate moment which starts from clip $i$ and ends at clip $j$, where $0\le i\le j\le N-1$.

Based on the 2D temporal map structure, we extract a feature map $\textbf{F}^M \in \mathbb{R}^{N \times N \times d^V}$ to represent candidate moments. Specifically, for each clip $v_i$ in the video, we extract features for its frames using a pretrained CNN model. The max-pooling result of frame features are taken as the clip representation $\textbf{f}_i\in\mathbb{R}^{d^V}, i=0, 1, \cdots, N-1$.  
The clip features serve as basic units to build the feature map. For a candidate moment $m_{ij}$, we apply a stacked convolution~\cite{MAN} on clips features $\{\textbf{f}_i,\textbf{f}_{i+1},\cdots,\textbf{f}_j\}$, and obtain $\textbf{F}_{ij}\in\mathbb{R}^{d^V}$ to represent $m_{ij}$ on the feature map $\textbf{F}^M$.

\subsection{Cross-Modal Matching Classifier}
The weakly supervised temporal language grounding problem can be tackled using a Multiple Instance Learning~(MIL) paradigm. In MIL, a set of bags are given, where each bag contains a collection of instances. For a specific bag, it is positive if at least one instance in the bag is positive, and it is negative if all instances in the bag are negative. 
Treating all possible video segments $m_{ij}$ as instances and the candidate set as the bag, the temporal grounding problem is natural to be formulated as an MIL problem, especially under video-level supervision. To learn learn an instance classifier for candidate moments with only bag labels, a video-level cross-modal matching classifier is necessary.

To learn cross-modal semantic alignment, we exploit 2D temporal adjacent network (2D-TAN) in a multiple instance learning manner, \emph{i.e.}, we build a cross-modal matching classifier based on 2D-TAN~\cite{2D-TAN}.
Then, to fuse features of two different modalities: $\textbf{h}_k \in \mathbb{R}^{d^S}$ for $k^{th}$ description sentence and $\textbf{F}_{ij} \in \mathbb{R}^{d^V}$ for the $(i,j)^{th}$ video segment on $\textbf{F}^M$, we adopt Hadamard product:
\begin{equation}
\textbf{F}_{ij,k} = \textbf{w}^S{\textbf{h}_k}^T \odot \textbf{w}^V{\textbf{F}_{ij}}^T,
\end{equation}
where $\textbf{w}^S \in \mathbb{R}^{d^F \times d^S}$ and $\textbf{w}^V \in \mathbb{R}^{d^F \times d^V}$ are learnable parameters of two fully-connected layers, $\cdot^T$ represents transpose of the vector, and $\odot$ represents Hadamard product. 
Consisting of $\textbf{F}_{ij,k}, i,j\in\{1, 2, \cdots, N\}$, the resulting feature map $\textbf{F}_k^M \in \mathbb{R}^{N \times N \times d^F}$ represents the relationship between all video segments and the $k^{th}$ input sentence.
However, the operations are performed separately on candidate moments, neglecting their temporal dependencies. To model context of each candidate moment, we exploit a temporal adjacent network on $\textbf{F}_k^M$, which consists of $L$ convolutional layers with same kernel size of $K$, and obtain a feature map $\textbf{F}^{M,c}_k$.	
The context-aware feature map $\textbf{F}^{M,c}_k$ are then fed into a prediction layer containing a fully-connected layer and a sigmoid function to get a 2D score map $\textbf{P}_k^M \in \mathbb{R}^{N \times N}$. The $(i,j)^{th}$ score on $\textbf{P}_k^M$ indicates the possibility of candidate moment $m_{ij}$ that corresponds to the input query. Note that the lower triangular part of $\textbf{P}_k^M$ is set to zero.

Intuitively, given a description sentence, the element giving max response value in the feature map is supposed to be the corresponding moment, and therefore, can be seen as a measurement for video-sentence alignment. Hence, we adopt a max operation on the score map $\textbf{P}_k^M$, \emph{i.e.},
\begin{equation}
P_k=\mathop{max}\limits_{i,j} \textbf{P}_k^M,
\end{equation}
and obtain the matching score $P_k$ for sentence $S_k$ and the input video.
As discussed above, we aggregate scores of all sentences in the input paragraph by a max operation, and take the result of most relevant and discriminative description as the final matching score, \emph{i.e.},
\begin{equation}
P=\mathop{max}\limits_{k} P_k,
\end{equation}
where $k\in \{0,1,\cdots,n_p-1\}$ represents the serial number of input sentences. 

We train the cross-modal classifier via a cross-entropy loss. 
Formally, we define matching label $y_m$, where $y_m=1$ indicates  the input paragraph is the corresponding description of the video, and 0 otherwise. The multiple instance learning loss is written as:
\begin{equation}
\mathcal{L}^{mil} = -(y_m\log P + (1-y_m)\log (1-P)),
\end{equation}
where $P$ is the score predicted by the cross-modal matching classifier.

\subsection{Self-Discriminating Loss}

\begin{figure}[htb]
	\centering
	\includegraphics[scale=0.55]{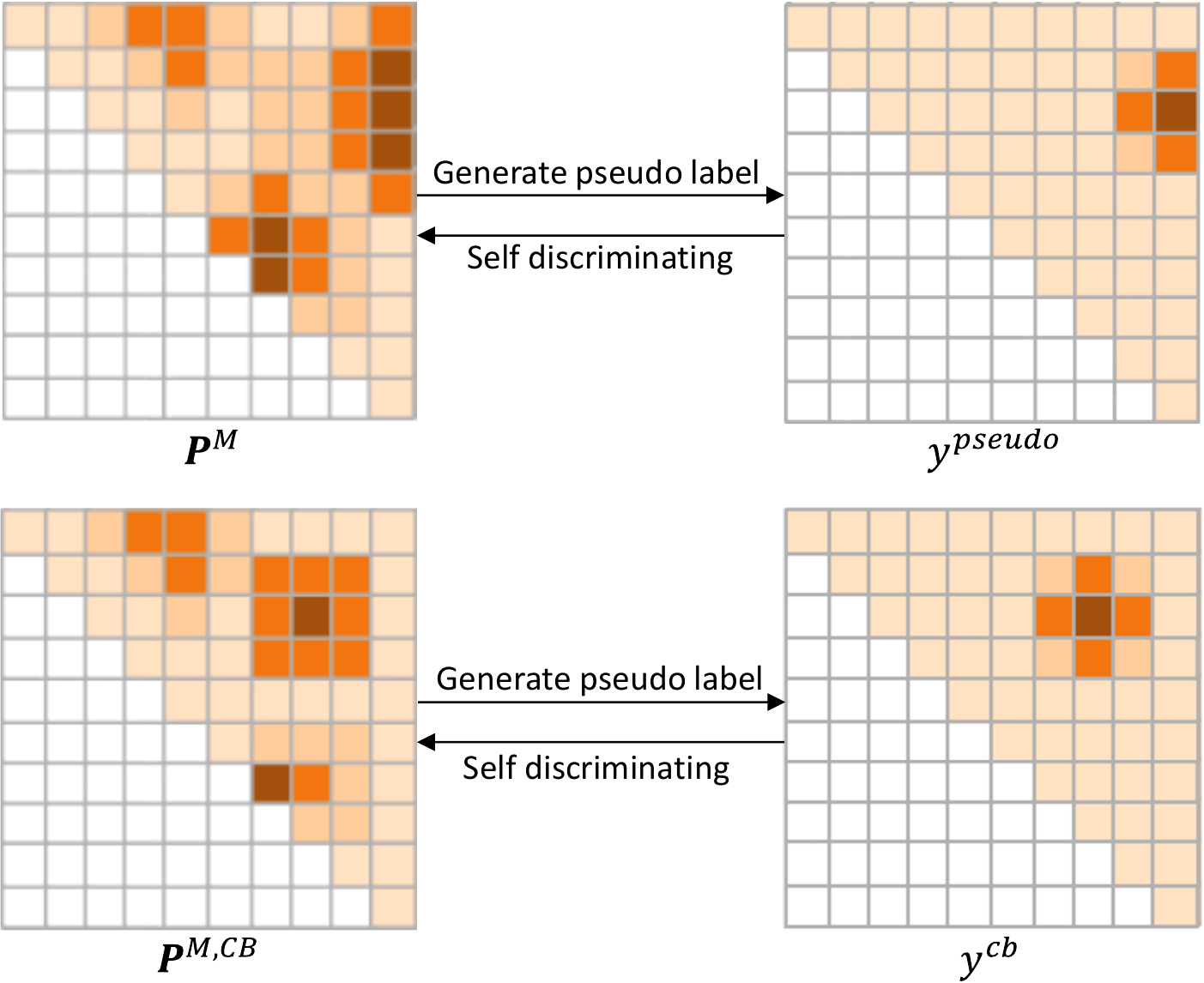}
	\caption{Illustration of the self-discriminating loss, leveraged on the cross-modal matching classifier and the complementary branch, respectively.}
	\label{fig:sdloss}
\end{figure}

During the training of the matching classifier introduced above, the cross-modal alignment of semantically meaningful contents can be learned. However, it is sufficient for the classification task to just response to semantic contents without distinguishing clips that contain them. Therefore, predicted score map is not discriminative enough to localize all contents in the language query, resulting in low precision of the predicted temporal boundaries.
In CIAN~\cite{CIAN}, a completion loss is introduced to compensate the sparsity of the CAM initialization for weakly supervised semantic segmentation. 
Inspired by it, we introduce self-discriminating loss to improve the precision of the output of WSTAN. 
As shown in Fig.~\ref{fig:sdloss}, the self-discriminating loss aims to distinguish the peak response on the 2D map and suppress others. Thus, the model is forced to be more temporally discriminative, improving the precision of prediction while acceptably decreasing the recall of foreground frames. 

Given the input sentence $S_k$, a 2D score map $\textbf{P}_k^M$ is predicted by the cross-modal matching classifier. We perform an $argmax$ operation on $\textbf{P}_k^M$ to get the most relevant video segment $m_{k,l}$, \emph{i.e.}, $m_{k,l}=argmax(\textbf{P}_k^M)$, where $l$ is the sequence number of the segment.
$m_{k,l}$ is then regarded as pseudo supervision to train an additional grounding branch.
Different from the hard labelling in~\cite{CIAN}, we devise a soft pseudo label generation strategy. 
Specifically, for each candidate moment $m_j=[s_j, e_j]$ on the 2D map, we compute the intersection-over-union (IoU) score $o_{k,j}$ with $m_{k,l}$, \emph{i.e.}, 
\begin{equation}
\begin{aligned}
o_{k,j}&=IoU(m_{k,l}, m_j) \\
	   &=\frac{\max(0,\min(e_{k,l}, e_j) - \max(s_{k,l}, s_j))}
	   {\max(e_{k,l}, e_j) - \min(s_{k,l}, s_j)}.
\end{aligned}
\end{equation}
To avoid influence of uncertainty introduced by pseudo supervision, we truncate the scores $o_{k,j}$ with thresholds $o_{min}$ and $o_{max}$.
The pseudo labels for self-training is written as:
\begin{equation}
y_{k,j}^{pseudo} = \left\{ 
\begin{array}{lr}
0, & o_{k,j}\leq o_{min} \\
\frac{o_{k,j}-o_{min}}{o_{max}-o_{min}}, & o_{min} < o_{k,j} < o_{max} \\
1, & o_{k,j}\geq o_{max} 
\end{array} 
\right..
\label{iou_label}
\end{equation}
Then we define the self-discriminating loss for each description sentence as:
\begin{equation}
\mathcal{L}^{sd}_k = \frac{1}{C}\sum_{j=1}^{C}y_{k,j}^{pseudo}\log p^j_k,
\label{loss_sd_0}
\end{equation}
where $p^j_k$ is the probability of $j^{th}$ candidate moment predicted in $\textbf{P}^M_k$.

In practice, because of the uncertainty of acquired pseudo labels, especially at the beginning of training, we reformulate Eq.~\eqref{loss_sd_0} into a weighted version:
\begin{equation}
\mathcal{L}^{sd}_k = \frac{w_k}{C}\sum_{j=1}^{C}y_{k,j}^{pseudo}\log p^j_k,
\end{equation}
where $w_k$ is the loss weight and is defined as $max(\textbf{P}_k^M)$. At the beginning of training, $w_k$ is relatively small, hence the $L^{sd}$ is small. Therefore, the noisy pseudo labels will not harm the performance of WSTAN dramatically.
During later training procedure, the model can achieve higher scores for ground truth moments, \emph{i.e.}, the model becomes more confident about its predictions, making the loss weight $w_k$ grow larger.

The overall self-discriminating loss is averaged among input description paragraph, \emph{i.e.}, 
\begin{equation}
\mathcal{L}^{sd} = \frac{1}{n_p}\sum_{k=0}^{n_p-1}\mathcal{L}^{sd}_k.
\end{equation}

\subsection{Complementary Branch}
Cross-modal semantic alignment learning is a straightforward idea to learn temporal relevance without explicit annotations. However, with the training objective of classification, the network is sensitive to semantically distinct objects, while ignoring some concepts mentioned in text queries. Therefore, clips that contain less or abstract semantic concepts, \emph{e.g.}, actions, postures, and non-rigid objects, which contribute less to the matching classification, are more likely to be classified as background. 
To improve the recall of semantic concepts in videos, we design a complementary branch to refine the predictions, and rediscover more semantically meaningful clips.
Specifically, we apply an additional prediction layer consisting of a fully-connected layer and a sigmoid function to $\textbf{F}^{M,c}_k$, the previously produced text-fused and context-aware feature map, and generate a score map $\textbf{P}^{M,CB}_k$.

The output of the complementary branch is supervised by the intermediate result $\textbf{P}^M_k$ of the cross-modal matching classifier.
Different from the self-discriminating loss, the pseudo supervision is used to supervise the output of another branch, and the constraint does not emphasize but refines the output head to improve semantic recall.
Formally, given the input sentence $S_k$, We get the most relevant moment $m_{k,l}=argmax(\textbf{P}_k^M)$. We take a similar soft label strategy as the self-discriminating loss to produce pseudo labels $y_{k,j}^{cb}$ based on $m_{k,l}$.
Then we define the complementary loss as follows,
\begin{equation}
\mathcal{L}^{cb}_k = \frac{1}{C}\sum_{j=1}^{C}y_{k,j}^{cb}\log p^j_k,
\label{loss_cb_0}
\end{equation}
where $p^j_k$ is the probability of $j^{th}$ candidate moment predicted in $\textbf{P}^{M,CB}_k$. Different from Eq.~\eqref{loss_sd_0}, where both $p_k^j$ and $y_{k,j}^{pseudo}$ are generated from $\textbf{P}_k^M$, the probabilities and labels are generated from predictions of different branches for $\mathcal{L}^{cb}_k$.

We also reformulate Eq.~\eqref{loss_cb_0} into a weighted manner,
and take the paragraph average as the overall complementary loss, \emph{i.e.}, 
\begin{equation}
\mathcal{L}^{cb} = \frac{1}{n_p}\sum_{k=0}^{n_p-1}\frac{w_k}{C}\sum_{j=1}^{C}y_{k,j}^{cb}\log p^j_k,
\end{equation}
where $w_k$ is the loss weight and is defined as $max(\textbf{P}_k^M)$.

With similar motivation to the alignment learning stage, we also deploy self-discriminating loss on the output $\textbf{P}^{M,CB}_k, k\in \{0,1,\cdots,n_p-1\}$ from complementary branch, denoted by $\mathcal{L}^{cb,sd}$,
\emph{i.e.},
\begin{equation}
\mathcal{L}^{cb,sd} = \frac{1}{n_p}\sum_{k=0}^{n_p-1}\frac{w_k}{C}\sum_{j=1}^{C}y_{k,j}^{cb,pseudo}\log p^j_k,
\end{equation}
where $y_{k,j}^{cb,pseudo}$ is produced based on the prediction of the complementary branch $\textbf{P}_{k}^{M,CB}$.

\subsection{Training Objective}

To train the WSTAN, as well as utilizing self-discriminating strategy, we take the weighted sum of MIL loss, self-discriminating loss, and complementary loss as the overall training objective. 
The overall training loss is written as:
\begin{equation}
\mathcal{L} = \alpha\mathcal{L}^{mil} + \beta\mathcal{L}^{cb} + \gamma\mathcal{L}^{sd} + \gamma\mathcal{L}^{cb,sd},
\end{equation}
where $\alpha, \beta, \gamma$ are hyperparameters and $\alpha+\beta+\gamma=1$. Note that the self-discriminating loss and complementary loss are added only when the input of two modalities are matched.

\subsection{Online Inference}
During inference, the network takes a sentence-video pair as input with only one sentence in the text input. WSTAN localizes the sentence by taking the index of the max response on the last predicted score map.
Specificity, if the complementary branch is adopted, the last score map is $\textbf{P}^{M,CB}$, and $m_{ij}^{CB} = argmax(\textbf{P}^{M,CB})$, \emph{i.e.}, the video segment from the $i^{th}$ clip to the $j^{th}$ clip is predicted as the best grounding result.
Otherwise, the prediction of the cross-modal matching classifier $\textbf{P}^M$ is taken as the last score map, and $m_{ij} = argmax(\textbf{P}^{M})$ is the inference result.

\section{Experiments}\label{experiments}

\subsection{Datasets}
We evaluate our approach on three benchmark datasets, \emph{i.e.}, ActivityNet-Captions, Charades-STA, and DiDeMo. 

\textbf{ActivityNet-Captions}
is a large-scale dataset with dense video captions, including 20k videos and over 70k moment-sentence pair annotations. Compared to the other two datasets, videos in the ActivityNet-Captions dataset are open-domain and much longer with an average duration of 117.74 seconds. 
The released ActivityNet-Captions dataset contains 17031 moment-description pairs for training, and 17505, 17031 in val\_1 and val\_2 sets. Following SCN~\cite{SCN}, we use val\_1 as validation set and val\_2 as test set.

\textbf{Charades-STA}
is introduced~\cite{TALL} for the temporal language localization task. Videos in the Charades dataset~\cite{Charades} are mainly about indoor everyday activities, with an average length of 29.8 seconds. As the original Charades dataset has only temporal activity localization and video-level paragraph description annotations, labels for temporal language grounding are added in Charades-STA~\cite{TALL} later. Charades-STA dataset contains 12408 moment-sentence pairs in the training set and 3720 pairs in the test set.

\textbf{DiDeMo} 
contains over 10k videos selected from Flickr, with over 40k temporally localized text descriptions. All the videos are trimmed to a maximum of 30 seconds and equally divided into six 5-second segments, while short videos with length lower than 25 seconds are aborted. Therefore, the temporal language grounding task can be seen as a candidate moment ranking problem, and for each video there are only 21 possible moment candidates. Videos in the DiDeMo dataset are randomly split into training, validation and test set containing 8395, 1065, and 1004 videos, respectively.
For each video, at least 4 annotators are assigned to label text description boundaries, hence the evaluation method on DiDeMo is slightly different from other datasets, as illustrated in the next subsection.

\begin{table}[tb]
	\centering
	\begin{threeparttable}
		\caption{Performance comparison on ActivityNet-Captions dataset.}
		\setlength{\tabcolsep}{1pt}{
			\begin{tabular}{ll|ccc|ccc}
				\toprule[2pt]
				\multirow{2}*{Method} & \multirow{2}*{Setting} & \multicolumn{3}{c|}{R@1} & \multicolumn{3}{c}{R@5} \\
				& & IoU=0.1 & IoU=0.3 & IoU=0.5 & IoU=0.1 & IoU=0.3 & IoU=0.5 \\
				\hline
				\hline
				2D-TAN~\cite{2D-TAN} & FS & - & 58.75 & 44.05 & - & 85.65 & 76.65 \\
				\hline
				Random & - & 38.23 &  18.64 &  7.63 & 75.74 & 52.78 & 24.49 \\
				WSLLN~\cite{WSLLN} & WS & 75.4 & 42.8 & 22.7 & - & - & - \\
				SCN~\cite{SCN} & WS & 71.48 & 47.23 & \underline{29.22}& 90.88 & 71.45 & \underline{55.69} \\
				\hline
				\textbf{WSTAN} Base & WS & \textbf{80.61} & 50.67 & 27.14 & \underline{91.57} & \underline{74.90} & 47.28 \\
				\textbf{WSTAN} Full & WS & \underline{79.78} & \textbf{52.45} & \textbf{30.01} & \textbf{93.15} & \textbf{79.38} & \textbf{63.42} \\
				\bottomrule[2pt]
			\end{tabular}
		}
		\label{res_act}
		\begin{tablenotes}
			\item[1] `FS' and `WS' means ``Fully Supervised'' and ``Weakly Supervised'', respectively.
			`Base' represent WSTAN with only cross-modal matching classifier, and `Full' represents WSTAN with self-discriminating loss and the complementary branch. 
			\item[2] The best and second best numbers are in \textbf{bold} and \underline{underlined}, respectively.
		\end{tablenotes}
	\end{threeparttable} 
\end{table}

\begin{table}[tb]
	\centering
	\begin{threeparttable}
	\caption{Performance comparison on Charades-STA dataset.}
	\setlength{\tabcolsep}{1pt}{
			\begin{tabular}{ll|ccc|ccc}
				\toprule[2pt]
				\multirow{2}*{Method} & \multirow{2}*{Setting} & \multicolumn{3}{c|}{R@1} & \multicolumn{3}{c}{R@5} \\
				& & IoU=0.3 & IoU=0.5 & IoU=0.7 & IoU=0.3 & IoU=0.5 & IoU=0.7 \\
				\hline
				\hline
				2D-TAN~\cite{2D-TAN}\ & FS & - & 39.81 & 23.25 & - & 79.33 & 52.15 \\
				\hline
				Random & - & - &  8.51 &  3.03 & - & 37.12 & 14.06 \\
				TGA~\cite{TGA} & WS & 29.68 & 17.04 & 6.93 & 83.87 & 58.17 & 26.80 \\
				SCN~\cite{SCN} & WS & 42.96 & \underline{23.58} & \underline{9.97} & \textbf{95.56} & \underline{71.80} & \underline{38.87} \\
				\hline
				\textbf{WSTAN} Base & WS & \textbf{51.61} & 21.64 & 9.01 & 89.76 & 66.24 & 37.07 \\
				\textbf{WSTAN} Full & WS & \underline{43.39} & \textbf{29.35} & \textbf{12.28} & \underline{93.04} & \textbf{76.13} & \textbf{41.53} \\
				\bottomrule[2pt]
			\end{tabular}
	}
	\label{res_cha}
	 \begin{tablenotes}
		\item[1] The best and second best numbers are in \textbf{bold} and \underline{underlined}, respectively.
	\end{tablenotes}
	\end{threeparttable} 
\end{table}

\subsection{Implementation Details}
\subsubsection{Evaluation Metrics}
We adopt different evaluation methods for the three datasets. 
Following~\cite{TALL}, On ActivityNet-Captions dataset, similarly, we report results for IoU$\in$\{0.5,0.3,0.1\} and Recall@\{1,5\}. 
On Charades-STA dataset, we report results for intersection-over-union (IoU)$\in$\{0.7,0.5,0.3\} and Recall@\{1,5\}. 
On the DiDeMo dataset, considering the limited number of candidates and labels from different annotators, we follow~\cite{DiDeMo}, and measure the performance of models with metrics: Rank@1, Rank@5, and mean intersection over union (mIoU). Here Rank@$k$ means the ground truth temporal boundaries labeled by different annotators are on average ranked higher than $k$ in prediction. Note that following~\cite{DiDeMo}, we only include the best-matched three ground truth labels, to reduce the influence of outliers.

\subsubsection{Experiment settings} \label{Evaluation_Metrics}
We utilize released visual features for all three datasets. Specifically, following prior works~\cite{2D-TAN,TGA,SCN}, we adopt VGG feature~\cite{VGG} for Charades-STA and DiDeMo, and C3D feature~\cite{C3D} for ActivityNet-Captions. The VGG features are 4096-dimensional, and we reduce the dimension of the C3D features from 4096 to 500 using PCA.
We set the number of sampled clips in a video, \emph{i.e.} $N$, to 16 for Charades-STA, 64 for ActivityNet-Captions, and 6 for DiDeMo, considering their video length and annotation characteristics. 
For temporal adjacent network, we adopt an 8-layer convolutional network with kernel size of 5 for Charades-STA and DiDeMo, and a 4-layer convolution network with kernel size of 9 for ActivityNet-Captions. We adopt a three-layer LSTM for language encoding.
The cross-modal negative sampling probability is set to 0.5. 
We apply non-maximum suppression (NMS) while inferencing.
An Adam optimizer is used during training and the learning rate is set to $1\times 10^{-4}$. 
Our code is available at https://github.com/ycWang9725/WSTAN.

\subsection{Results and Analysis}

\begin{figure}[tb]
	\begin{minipage}[b]{0.82\linewidth}
		\centering
		\subfloat[]{
			\includegraphics[scale=0.46]{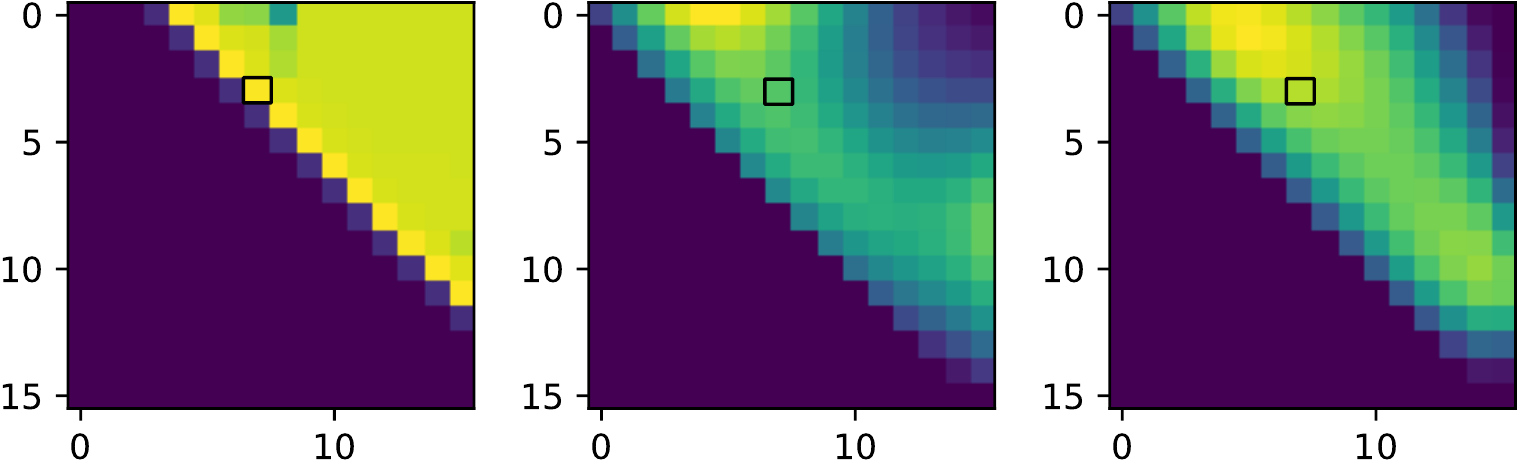}
		}
		
		\subfloat[]{
			\includegraphics[scale=0.46]{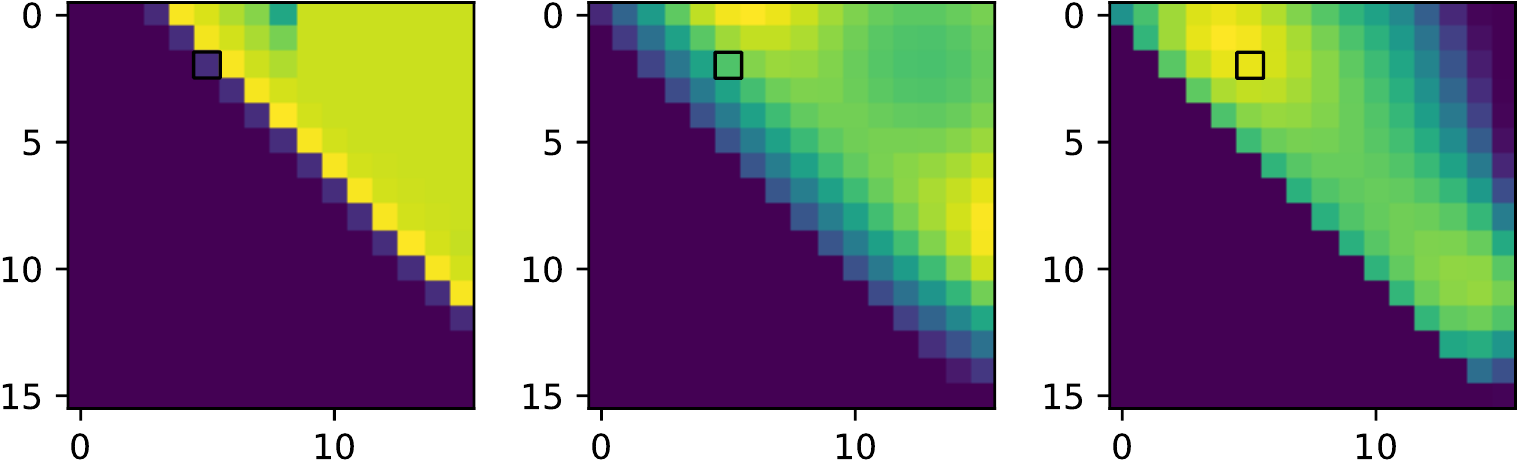}
		}
		
		\subfloat[]{
			\includegraphics[scale=0.46]{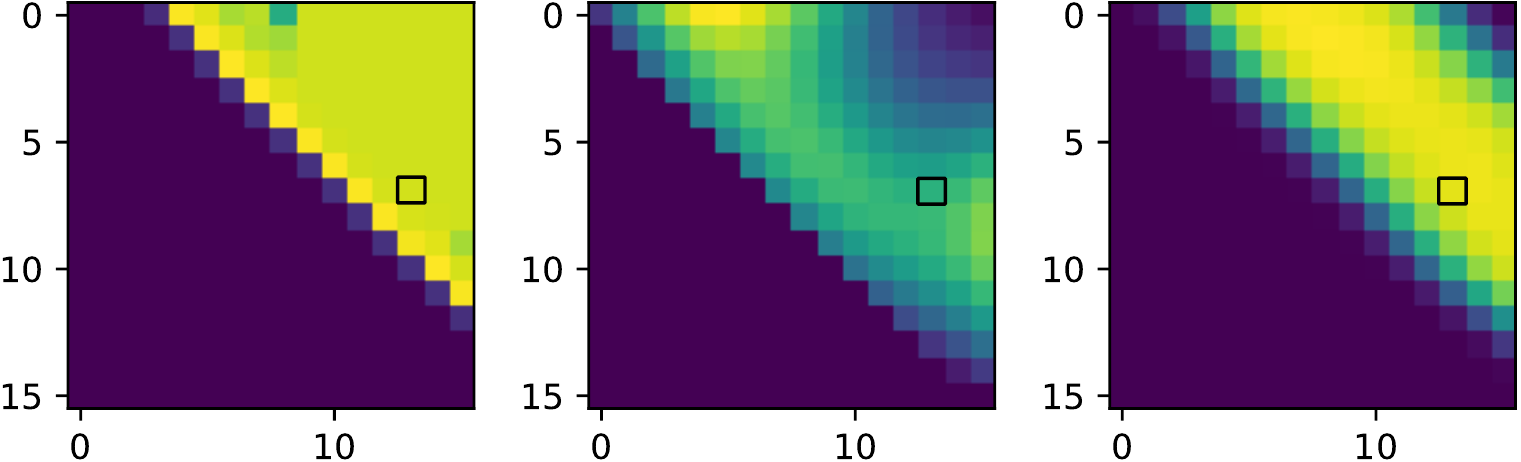}
		}
	\end{minipage} 
	\hfill
	\begin{minipage}[b]{0.15\linewidth}
		\centering
		\subfloat{
			\includegraphics[scale=0.69]{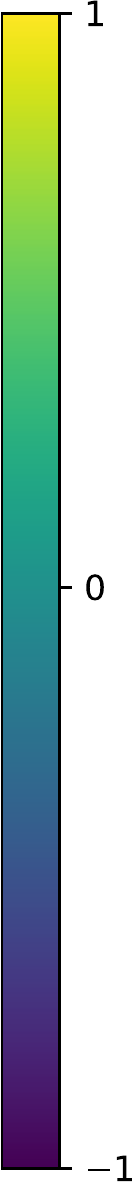}
		}
	\end{minipage}
	
	\caption{Visualization of score maps predicted by models trained under different settings.
		From left to right: `Base' model with either self-discriminating loss or complementary branch, `Base' model with self-discriminating loss, and `Full' Model. Black boxes indicate ground truth video segment.
	}
	\label{fig:heatmaps}
\end{figure}

We compare our WSTAN with recently proposed state-of-the-art methods, including both \textit{weakly supervised} methods: TGA~\cite{TGA} on Charades-STA and DiDeMo, SCN~\cite{SCN} on Charades-STA and ActivityNet-Captions, WSLLN~\cite{WSLLN} on DiDeMo and ActivityNet-Captions;
and \textit{fully supervised} methods: TGN~\cite{TGN} on DiDeMo, 2D-TAN~\cite{2D-TAN} on all three datasets.
Also, we report a random baseline, which means the output of the model is randomly selected from all possible moment candidates.

Experiment results of our proposed WSTAN on three datasets are summarized in Table~\ref{res_cha} for Charades-STA, Table~\ref{res_act} for ActivityNet-Captions, and Table~\ref{res_did} for DiDeMo.
By observing the evaluation results, we can see that WSTAN with self-discriminating loss and complementary branch (the `Full' setting) outperforms state-of-the-art weakly supervised methods on all three benchmark datasets, and achieves comparable performance with some state-of-the-art fully supervised methods. 

\begin{table}[tb]
	\centering
	\begin{threeparttable}
		\caption{Performance comparison on DiDeMo dataset.}
		\setlength{\tabcolsep}{10pt}{
			\begin{tabular}{ll|ccc}
				\toprule[2pt]
				Method & Setting & R@1 &R@5 & mIoU \\
				\hline
				\hline
				TGN~\cite{TGN} & FS & 28.23 &79.26 & 42.97 \\
				2D-TAN~\cite{2D-TAN} & FS & 19.40 & 68.64 & 31.92 \\
				\hline
				Random & - & 3.75 & 22.5 & 22.64 \\
				TGA~\cite{TGA} & WS & 12.19 & 39.74 & 24.92 \\
				WSLLN~\cite{WSLLN} & WS & \textbf{19.40} & \underline{53.10} & 25.40 \\
				\hline
				\textbf{WSTAN} Base & WS & \textbf{19.40} & 45.59 & \underline{31.92} \\
				\textbf{WSTAN} Full & WS & \textbf{19.40} & \textbf{54.64} & \textbf{31.94} \\
				\bottomrule[2pt]
			\end{tabular}
		}
		\label{res_did}
		\begin{tablenotes}
			\item[1] The best and second best numbers are in \textbf{bold} and \underline{underlined}, respectively.
			\item[2] We evaluated 2D-TAN~\cite{2D-TAN} on DiDeMo for comparison, as the authors did not report their results.
		\end{tablenotes}
	\end{threeparttable}
\end{table}

To be specific, the results lead to the following facts:
\begin{itemize}
	\item Compared with random baseline, the performance of our method has a significant improvement. Thus our assumption is validated that the weakly supervised temporal language grounding problem can be formulated as an MIL problem, and the training objective of the WSTAN, \emph{i.e.}, cross-modal alignment learning, is straightforward yet effective.
	
	\item Compared with TGA~\cite{TGA}: TGA uses a sliding window to select candidate moments and attempts to learn cross-modal alignment by mapping visual and textual input into a unified latent space. Our method can obtain better results even without self-discriminating loss or the complementary branch, showing the effectiveness of the temporal adjacent network with paragraph text input, as well as our strategy to sample proposal candidates. 
	
	\item Compared with SCN~\cite{SCN}: SCN proposes a candidate proposal scoring model learned by sentence completion. Performance of our base WSTAN is similar to SCN. However, after adding self-discriminating loss and the complementary branch, our method outperforms SCN on almost all metrics by a large margin, especially on ActivityNet-Captions dataset. 
	The sentence completion learning objective of SCN is similar to our motivation to learn semantic alignment more distinctly. However, our model can refine the predictions and learn more semantic concepts, leading to better performance.
	
	\item Compared with WSLLN~\cite{WSLLN}: WSLLN proposed a two-branch framework that measures segment-text consistency and conducts segmentation selection simultaneously. Our method outperforms WSLLN overall, which also indicates the effectiveness of our self-discriminating loss and the complementary branch. Note that the performance of WSLLN on ActivityNet-Captions dataset is only reported at R@1.
\end{itemize}

	The motivation of the proposed method is to improve temporal precision without explicit boundaries for supervision. Therefore, the improvement of WSTAN is more notable on strict metrics~(higher IoU requirements), which indicates the ability of WSTAN to predict temporal boundaries precisely. Differently, with semantic complementing as the training objective, SCN manages to better learn cross-modal semantic alignment on word-level, which helps to localize difficult descriptions in video, and therefore boost the number on metrics with lower IoU requirement.

It can be observed that the R@1 performance of WSTAN on DiDeMo dataset is not improved from `Base' to `Full', and even equates the performance of the fully-supervised 2D-TAN. The limitation is due to the characteristics of the DiDeMo dataset.
As discussed above, the R@1 metric for DiDeMo measures the top-1 accuracy of the selection from fixed number~(21) of candidates. The short average duration of videos and small number of clips restricts the temporal adjacent network to learn among candidates. 
By observing the model outputs, we find that the model tends to predict one-clip moment, which may be the origin of the number~(19.40\%).
However, these reasons not only affect TAN-based methods, but also other frameworks, like WSLLN in Table III.

\subsection{Ablation Study}

In this section, we analyse the effectiveness of different parts in WSTAN by performing ablation experiments. Results are shown in Table~\ref{ablation} and Table~\ref{ablation_input}. 
We also visualize some results inFig.~\ref{fig:heatmaps} and Fig.~\ref{fig:results} to validate the effect of different parts of WSTAN qualitatively. 

\begin{table}[tb]
	\centering
	\begin{threeparttable}
		\caption{
			Influence of different parts of WSTAN.
		}
		\setlength{\tabcolsep}{1.2pt}{
			\begin{tabular}{ccc|ccc|ccc}
				\toprule[2pt]
				\multirow{2}*{SD@MIL} & \multirow{2}*{CB} & \multirow{2}*{SD@CB} & \multicolumn{3}{c|}{R@1} & \multicolumn{3}{c}{R@5} \\
				&&& IoU=0.3 & IoU=0.5 & IoU=0.7 & IoU=0.3 & IoU=0.5 & IoU=0.7 \\
				\hline
				\hline
				\XSolidBrush & \XSolidBrush & \XSolidBrush & \textbf{51.61} & 21.64 & 9.01 & 89.76 & 66.24 & 37.07 \\
				& \CheckmarkBold &  & 41.94 & 21.45 & 8.17 & 71.72 & 49.25 & 19.41 \\
				& \CheckmarkBold & \CheckmarkBold & 40.19 & 25.75 & 9.25 & 78.39 & 47.88 & 20.70 \\
				\CheckmarkBold &  &  & 42.23 & 27.85 & \underline{13.66} & \textbf{95.19} & \underline{68.82} & 35.32 \\
				\CheckmarkBold & \CheckmarkBold &  & 39.65 & \textbf{29.73} & \textbf{14.70} & 91.59 & 68.55 & \underline{37.77} \\
				\CheckmarkBold & \CheckmarkBold & \CheckmarkBold & \underline{43.39} & \underline{29.35} & 12.28 & \underline{93.04} & \textbf{76.13} & \textbf{41.53} \\
				\bottomrule[2pt]
			\end{tabular}
		}
		\label{ablation}
		\begin{tablenotes}
			\item[1] `SD' refers to self-discriminating loss and `CB' refers to complementary branch. ``SD@MIL'' and ``SD@CB'' refers to apply SD loss to the cross-modal matching classifier and to the complementary branch, respectively.
			\item[2] The first row corresponds to the `Base' setting  and the last row  corresponds to the `Full' setting above.
			\item[3] The performance is reported on Charades-STA dataset.
		\end{tablenotes}
	\end{threeparttable}
\end{table}

\begin{table}[tb]
	\centering
	\caption{Comparison among different type of text input on Charades-STA dataset.
}
	\setlength{\tabcolsep}{3pt}{
		\begin{tabular}{l|ccc|ccc}
			\toprule[2pt]
			\multirow{2}*{Input Type} & \multicolumn{3}{c|}{R@1} & \multicolumn{3}{c}{R@5} \\
			& IoU=0.3 & IoU=0.5 & IoU=0.7 & IoU=0.3 & IoU=0.5 & IoU=0.7 \\
			\hline
			\hline
			Paragraph & 43.39 & \textbf{29.35} & \textbf{12.28} & \textbf{93.04} & \textbf{76.13} & \textbf{41.53} \\
			Single & 49.38 & 21.88 & 7.90 & 91.80 & 64.62 & 32.74  \\
			Single (soft) & 49.46 & 25.59 & 9.97 & 90.86 & 65.13 & 35.30  \\
			Video Script & \textbf{50.70} & 26.16 & 11.05 & 91.91 & 62.66 & 31.85  \\
			\bottomrule[2pt]
		\end{tabular}
	}
	\label{ablation_input}
\end{table}

\subsubsection{Influence of Self-Discriminating Loss}
The score map predicted by WSTAN with and without self-discriminating loss is shown in Fig.~\ref{fig:heatmaps} (the left and middle figure for each example). It can be observed that SD loss improves the temporal discrimination ability markedly by suppressing the scores of adjacent candidates of the peak response.
We compare the performance of the WSTAN with and without self-discriminating loss under network settings, shown in Table~\ref{ablation}.
By adding self-discriminating loss, the performance is improved on Recall@1, IoU=0.5 and IoU=0.7, proving the enhanced precision of predicted temporal boundaries. However, numbers of other metrics drop slightly, proving the idea that the pseudo supervision introduced by the self-discriminating loss can damage the recall of foreground frames.

\subsubsection{Influence of Complementary Branch}
The score map predicted by WSTAN with and without complementary branch is shown in Fig.~\ref{fig:heatmaps} (the middle and right figure for each example). The self-supervision of self-discriminating loss tends to propose a long segment consists of almost half of all clips, while the additional branch gives more precise temporal boundaries. Therefore the ground truth segment is given better score, \emph{i.e.}, higher rank, after the uncovering process of the branch.
We add complementary branch to WSTAN to improve the capability of precisely localizing moments. However, temporal precision is often opposite to the temporal recall. That is, if we pursue higher precision for most cases, the recall of difficult cases will drop.
We compare performance of the complementary branch with and without self-discriminating loss, shown in Table~\ref{ablation}.
Because of the low discrimination ability of WSTAN without self-discriminating loss, adding the additional branch brings a huge performance drop, which is intuitive. 
However, when self-discriminating loss is deployed on the base classifier (last three rows in Table~\ref{ablation}), the overall performance is improved by adding the complementary branch. 
It can be observed that by adding the `CB', the performance of WSTAN is improved on the metrics with stricter requirements~(Recall@1, IoU=0.5/0.7), while on looser metrics~(R@5, or IoU=0.1), the performance drops. However, by adding a SD constraint on CB, both the two influence are restrained to a balance of precision and recall, which achieve good performance on all metrics.

\subsubsection{Influence of Paragraph Input}

\begin{table}[tb]
	\centering
	\begin{threeparttable}
		\caption{Influence of the lower threshold for pseudo supervision. The settings used in our final model are underlined.}
		\setlength{\tabcolsep}{3pt}{
			\begin{tabular}{cc|ccc|ccc}
				\toprule[2pt]
				\multirow{2}*{th@SD} & \multirow{2}*{th@CB} & \multicolumn{3}{c|}{R@1} & \multicolumn{3}{c}{R@5} \\
				& & IoU=0.3 & IoU=0.5 & IoU=0.7 & IoU=0.3 & IoU=0.5 & IoU=0.7 \\
				\hline
				\hline
				\underline{0.9} & \underline{0.9} & 43.49 & \textbf{29.35} & 12.28 & 93.04 & 76.13 & 41.53 \\
				0.9 & 1.0 & 37.02 & 29.19 & \textbf{13.87} & 88.47 & 73.15 & 39.41 \\
				0.9 & 0.5 & 41.40 & 27.04 & 11.05 & 94.44 & \textbf{77.63} & 40.46 \\
				0.9 & 0.0 & 43.66 & 28.92 & 12.34 & 85.94 & 62.07 & 27.88 \\
				1.0 & 0.9 & \textbf{49.44} & 28.76 & 13.17 & \textbf{95.32} & 68.66 & 36.94 \\
				0.5 & 0.9 & 30.91 & 19.81 & 10.59 & 88.31 & 66.02 & 41.05 \\
				0.0 & 0.9 & 30.94 & 19.78 & 10.56 & 80.75 & 63.20 & \textbf{45.75} \\
				\bottomrule[2pt]
			\end{tabular}
		}
		\label{ablation_th}
		\begin{tablenotes}
			\item[1] `SD' and `CB' refer to self-discriminating loss and complementary branch, respectively.
			\item[2] `th' refers to the lower threshold of the pseudo labels.
		\end{tablenotes}
	\end{threeparttable}
\end{table}

\begin{figure*}[tb]
	\centering
	\includegraphics[scale=0.255]{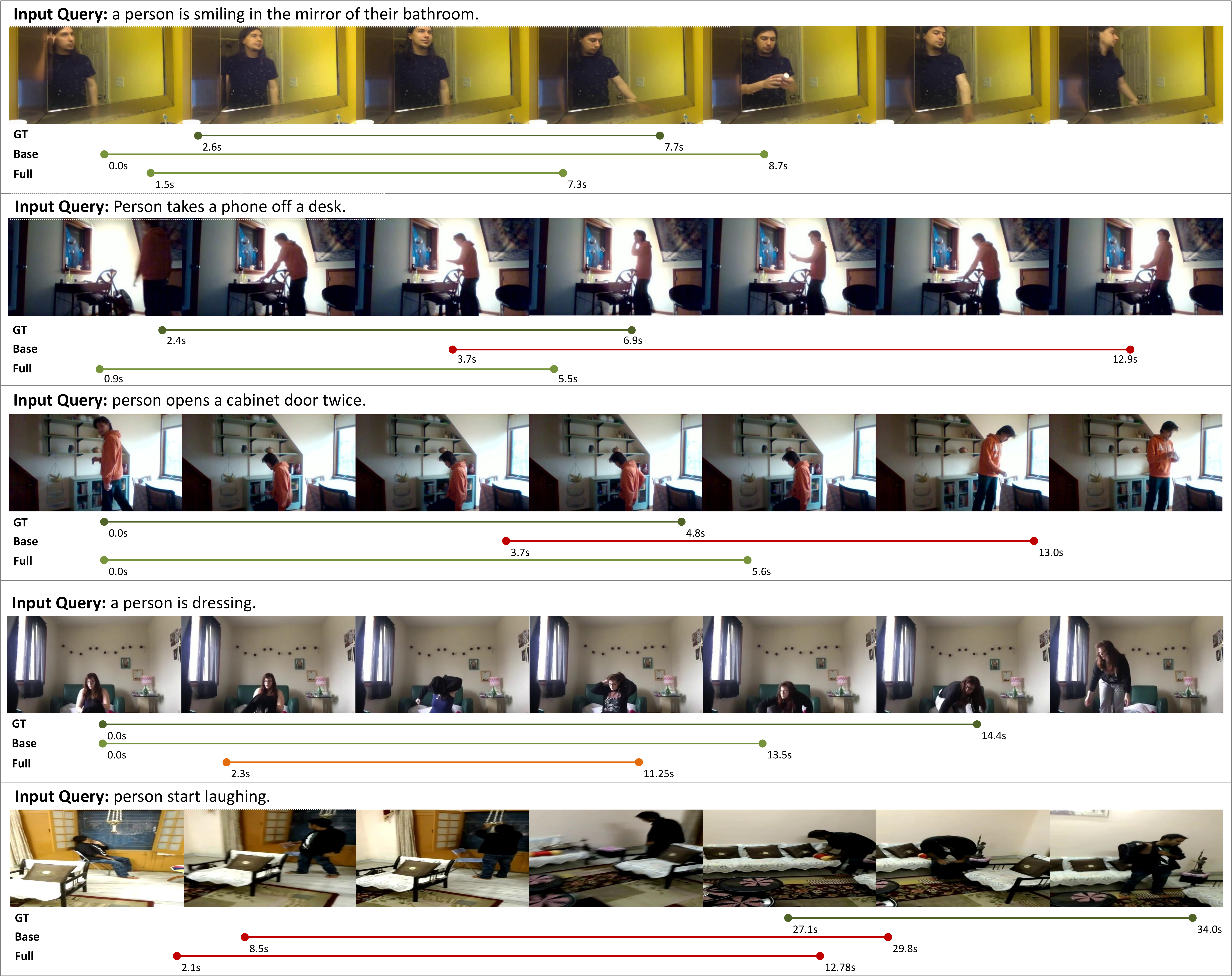}
	\caption{Success and failure examples on Charades-STA dataset. 
		(1) Both models localize the input query successfully, yet our `Full' model obtains better IoU (0.76) than the base model (0.59). 
		(2)(3) The `Base' model fails to localize keywords (`off', `open', `twice'), leading to failure, while the `Full' model acquires higher IoU. 
		(4) The `Base' model gets better IoU than the full model. However, by observing the video, we found the prediction of our `Full' model is even more precise for the 'dressing' action than ground truth, which also validates our opinion that temporal annotations can be inconsistent and ambiguous. 
		(5) Difficult case where both models fail to localize the language query.}
	\label{fig:results}
\end{figure*}

In order to validate the effectiveness of our paragraph text input design, we evaluate our method without aggregating the language descriptions of a video. 
As introduced above, we concatenate all the corresponding descriptions for a video, and train WSTAN with paragraph-video pairs. 
A max operation is performed on separately predicted scores to get the final prediction for cross-modal alignment learning. 
In the single query input setting, we use a sentence-video pair as input and omit the merge operation among predictions. 
To alleviate the disadvantage of hard matching, we also devised a soft matching strategy for single query input, where the matching label is the coverage rate of the words in the sampled query and the ground truth query.
Results (row 1$^{st}, 2^{nd}$ and row 3$^{th}$ in Table~\ref{ablation_input}) show that our paragraph input design can improve performance on most of the evaluation metrics, proving our intuition that the paragraph aggregating method can reduce the noise brought by semantically partial similarity among different videos. 
Due to the short length of videos in Charades-STA dataset, the requirement of IoU=0.3 is relatively weak for evaluation.
However, the performance drop on Recall@1, IoU=0.3 shows that the recall of foreground frames is slightly harmed by aggregating text input. 
In addition, the soft matching strategy indeed compensates for the single query setting, while still inferior to the paragraph input setting. Better soft matching strategy will be explored in the future.

\subsubsection{Influence of Pseudo Label Thresholds}
To alleviate the uncertainty of pseudo supervision, we truncate the pseudo labels with thresholds $o_{max}$ and $o_{min}$ during generation. 
In our experiments, the thresholds of pseudo labels are set to $o_{max}=1.0, o_{min}=0.9$ for both self-discriminating loss and the complementary branch. We conduct experiments to validate the impact of the lower thresholds $o_{min}$, and results are listed in Table~\ref{ablation_th}. It can be observed that the threshold of pseudo labels for self-discriminating loss mainly influences the performance on R@1, while the threshold for complementary branch mainly influences performance on R@5.

\subsection{Qualitative results}
Fig.~\ref{fig:results} shows some grounding results predicted by WSTAN. It can be observed that the full WSTAN is able to predict temporal boundaries more precisely, proving that cross-modal semantic alignment learning can be enhanced by adding complementary branch and leveraging self-discriminating loss. 
For example, in the second case in Fig.~\ref{fig:results}, the key object is `phone' and `desk', but `off' is also one of the most import semantic concept for predicting the temporal boundaries. The `Base' model has detected the small object `phone', which is useful in matching classification, while the keyword `off' is neglected. However, the `Full' model rediscover the keyword `off', and acquire higher IoU.

\subsection{Further Discussion} 
\subsubsection{Video-Level Description}
\label{script_explain}
As mentioned before, we tackle the weakly supervised temporal language grounding problem based on the assumption that video-level descriptions can be acquired with lower cost, and can therefore be extended to videos with more diverse content. However, existing methods focus on ready-made datasets and transfer to weakly supervision by simply omitting the temporal annotations. 
In our opinion, it is more common for videos to have video-level coarse descriptions with fewer sentences and more complex syntax, and we wonder the validity of WSTAN under this kind of supervision for training.

In the original Charades~\cite{Charades} dataset, the videos are recorded following a script, and single sentence descriptions are annotated afterwards. Both forms of description contain events that happen in videos, while the scripts are more difficult to analyse.
We evaluate our method on the video scripts and results (last row in Table~\ref{ablation}) show comparable performance are achieved by WSTAN with description sentences input. 
Although the video script are basically an aggregated version of description paragraph, we hope our attempt to be inspiring to explore temporal language grounding in more practical scenarios.

\subsubsection{Complementary Branch}
The idea of complementary branch is to improve the temporal precision with coarse video-level supervision, which is similar to the online refinement for weakly supervised object detection. In fact, the motivation is to generate pseudo labels to refine the MIL results using an additional branch. Therefore, the idea is applicable to weakly supervised spatial/temporal grounding problems, \emph{e.g.} spatial visual grounding, temporal action localization, and semantic segmentation. In addition, complementary branch can also be formed recurrently, where each branch refines result of the previous one iteratively. This iterative refinement mechanism has been proved effective on object detection, and can be further explored in other topics.

\section{Conclusion}
In this paper, we propose a novel weakly supervised temporal adjacent network (WSTAN) for the language grounding problem, which learns cross-modal semantic alignment in an MIL paradigm with a whole paragraph text input.
To enhance the temporal precision of localization, we further devise a self-discriminating loss and an additional complementary branch. 
We conducted experiments on three benchmark datasets, \emph{i.e.}, Charades-STA, ActivityNet-Captions, and DiDeMo.
Experimental results demonstrate the effectiveness of our approach. 
We also extend to a setting with weaker supervision, where only video-level coarse scripts are provided, and we validate our method under this supervision.


%

\appendices



\ifCLASSOPTIONcaptionsoff
  \newpage
\fi



%

\bibliographystyle{IEEEtran}
\bibliography{IEEEabrv,bib/refs}

\begin{thebibliography}{10}
\providecommand{\url}[1]{#1}
\csname url@samestyle\endcsname
\providecommand{\newblock}{\relax}
\providecommand{\bibinfo}[2]{#2}
\providecommand{\BIBentrySTDinterwordspacing}{\spaceskip=0pt\relax}
\providecommand{\BIBentryALTinterwordstretchfactor}{4}
\providecommand{\BIBentryALTinterwordspacing}{\spaceskip=\fontdimen2\font plus
\BIBentryALTinterwordstretchfactor\fontdimen3\font minus
  \fontdimen4\font\relax}
\providecommand{\BIBforeignlanguage}[2]{{%
\expandafter\ifx\csname l@#1\endcsname\relax
\typeout{** WARNING: IEEEtran.bst: No hyphenation pattern has been}%
\typeout{** loaded for the language `#1'. Using the pattern for}%
\typeout{** the default language instead.}%
\else
\language=\csname l@#1\endcsname
\fi
#2}}
\providecommand{\BIBdecl}{\relax}
\BIBdecl

\bibitem{VQA}
A.~Agrawal, J.~Lu, S.~Antol, M.~Mitchell, C.~L. Zitnick, D.~Parikh, and
  D.~Batra, ``Vqa: Visual question answering,'' \emph{Int. J. Comput. Vis.},
  vol. 123, pp. 4--31, 2015.

\bibitem{8988148}
J.~{Yu}, W.~{Zhang}, Y.~{Lu}, Z.~{Qin}, Y.~{Hu}, J.~{Tan}, and Q.~{Wu},
  ``Reasoning on the relation: Enhancing visual representation for visual
  question answering and cross-modal retrieval,'' \emph{IEEE Trans.
  Multimedia}, vol.~22, no.~12, pp. 3196--3209, 2020.

\bibitem{MovieQA}
M.~Tapaswi, Y.~Zhu, R.~Stiefelhagen, A.~Torralba, R.~Urtasun, and S.~Fidler,
  ``Movieqa: Understanding stories in movies through question-answering,''
  \emph{Proc. IEEE Conf. Comput. Vis. and Pattern Recog.}, pp. 4631--4640,
  2016.

\bibitem{8811730}
W.~{Zhang}, S.~{Tang}, Y.~{Cao}, S.~{Pu}, F.~{Wu}, and Y.~{Zhuang}, ``Frame
  augmented alternating attention network for video question answering,''
  \emph{IEEE Trans. Multimedia}, vol.~22, no.~4, pp. 1032--1041, 2020.

\bibitem{CLEVRAD}
J.~Johnson, B.~Hariharan, L.~van~der Maaten, L.~Fei-Fei, C.~L. Zitnick, and
  R.~B. Girshick, ``{CLEVR}: A diagnostic dataset for compositional language
  and elementary visual reasoning,'' \emph{Proc. IEEE Conf. Comput. Vis. and
  Pattern Recog.}, pp. 1988--1997, 2017.

\bibitem{COG}
G.~Yang, I.~Ganichev, X.~Wang, J.~Shlens, and D.~Sussillo, ``A dataset and
  architecture for visual reasoning with a working memory,'' \emph{Proc. Europ.
  Conf. Comp.}, p. 729–745, 2018.

\bibitem{7984828}
L.~{Gao}, Z.~{Guo}, H.~{Zhang}, X.~{Xu}, and H.~T. {Shen}, ``Video captioning
  with attention-based lstm and semantic consistency,'' \emph{IEEE Trans.
  Multimedia}, vol.~19, no.~9, pp. 2045--2055, 2017.

\bibitem{9121763}
Z.~{Zhang}, D.~{Xu}, W.~{Ouyang}, and L.~{Zhou}, ``Dense video captioning using
  graph-based sentence summarization,'' \emph{IEEE Trans. Multimedia}, pp.
  1--1, 2020.

\bibitem{8031355}
L.~Li, S.~Tang, Y.~Zhang, L.~Deng, and Q.~Tian, ``Gla: Global–local attention
  for image description,'' \emph{IEEE Transactions on Multimedia}, vol.~20,
  no.~3, pp. 726--737, 2018.

\bibitem{huangStorytelling}
T.-H.~K. Huang, F.~Ferraro, N.~Mostafazadeh, I.~Misra, A.~Agrawal, J.~Devlin,
  R.~Girshick, X.~He, P.~Kohli, D.~Batra, C.~L. Zitnick, D.~Parikh,
  L.~Vanderwende, M.~Galley, and M.~Mitchell, ``Visual storytelling,'' in
  \emph{Conf. North Am. Chapter Assoc. Comput. Linguist.: Hum. Lang. Technol.,
  NAACL HLT - Proc. Conf.}, 2016, pp. 1233--1239.

\bibitem{VideoStory}
S.~Gella, M.~Lewis, and M.~Rohrbach, ``A dataset for telling the stories of
  social media videos,'' in \emph{Proc. Conf. Empir. Methods Nat. Lang.
  Process., EMNLP}, 2018, pp. 968--974.

\bibitem{TALL}
J.~Gao, C.~Sun, Z.~Yang, and R.~Nevatia, ``Tall: Temporal activity localization
  via language query,'' in \emph{Proc. IEEE Int. Conf. Comput. Vis.}, 2017, pp.
  5277--5285.

\bibitem{DiDeMo}
L.~A. Hendricks, O.~Wang, E.~Shechtman, J.~Sivic, T.~Darrell, and B.~C.
  Russell, ``Localizing moments in video with natural language,'' in
  \emph{Proc. IEEE Int. Conf. Comput. Vis.}, 2017, pp. 5804--5813.

\bibitem{ijcai2018-143}
A.~Wu and Y.~Han, ``Multi-modal circulant fusion for video-to-language and
  backward,'' in \emph{Proc. Int. Joint Conf. Artif. Intell.}, 7 2018, pp.
  1029--1035.

\bibitem{Ge2019MACMA}
R.~Ge, J.~Gao, K.~Chen, and R.~Nevatia, ``Mac: Mining activity concepts for
  language-based temporal localization,'' in \emph{Proc. IEEE Winter Conf.
  Appl. Comput. Vis.}, 2019, pp. 245--253.

\bibitem{aaai2019Yuan}
Y.~Yuan, T.~Mei, and W.~Zhu, ``To find where you talk: Temporal sentence
  localization in video with attention based location regression,'' in
  \emph{Proc. AAAI Conf. Artif. Intell.}, 04 2018, pp. 9159--9166.

\bibitem{MAN}
D.~Zhang, X.~Dai, X.~Wang, Y.~Wang, and L.~Davis, ``Man: Moment alignment
  network for natural language moment retrieval via iterative graph
  adjustment,'' in \emph{Proc. IEEE Conf. Comput. Vis. and Pattern Recog.},
  2019, pp. 1247--1257.

\bibitem{2D-TAN}
S.~Zhang, H.~Peng, J.~Fu, and J.~Luo, ``Learning 2d temporal adjacent networks
  for moment localization with natural language,'' in \emph{Proc. AAAI Conf.
  Artif. Intell.}, 2020, p. 12870–12877.

\bibitem{He2019ReadWA}
D.~He, X.~Zhao, J.~Huang, F.~Li, X.~Liu, and S.~Wen, ``Read, watch, and move:
  Reinforcement learning for temporally grounding natural language descriptions
  in videos,'' in \emph{Proc. AAAI Conf. Artif. Intell.}, 2019, pp. 8393--8400.

\bibitem{cvprWangHW19}
W.~Wang, Y.~Huang, and L.~Wang, ``Language-driven temporal activity
  localization: {A} semantic matching reinforcement learning model,'' in
  \emph{Proc. IEEE Conf. Comput. Vis. and Pattern Recog.}, 2019, pp. 334--343.

\bibitem{TGA}
N.~C. Mithun, S.~Paul, and A.~K. Roy-Chowdhury, ``Weakly supervised video
  moment retrieval from text queries,'' in \emph{Proc. IEEE Conf. Comput. Vis.
  and Pattern Recog.}, June 2019, pp. 11\,584--11\,593.

\bibitem{WSLLN}
M.~Gao, L.~Davis, R.~Socher, and C.~Xiong, ``Wslln: Weakly supervised natural
  language localization networks,'' in \emph{Proc. Conf. Empir. Methods Nat.
  Lang. Process. Int. Jt. Conf. Nat. Lang. Process., Proc. Syst. Demonstr.},
  2019, pp. 1481--1487.

\bibitem{SCN}
Z.~Lin, Z.~Zhao, Z.~Zhang, Q.~Wang, and H.~Liu, ``Weakly-supervised video
  moment retrieval via semantic completion network,'' in \emph{Proc. AAAI Conf.
  Artif. Intell.}, Apr 2020, p. 11539–11546.

\bibitem{OICR}
P.~Tang, X.~Wang, X.~Bai, and W.~Liu, ``Multiple instance detection network
  with online instance classifier refinement,'' in \emph{Proc. IEEE Conf.
  Comput. Vis. and Pattern Recog.}, 2017, pp. 3059--3067.

\bibitem{ActivityNet}
R.~Krishna, K.~Hata, F.~Ren, L.~Fei-Fei, and J.~C. Niebles, ``Dense-captioning
  events in videos,'' in \emph{Proc. IEEE Int. Conf. Comput. Vis.}, 2017, pp.
  706--715.

\bibitem{ImageNet}
\BIBentryALTinterwordspacing
A.~Krizhevsky, I.~Sutskever, and G.~E. Hinton, ``Imagenet classification with
  deep convolutional neural networks,'' \emph{Commun. ACM}, vol.~60, no.~6, p.
  84–90, May 2017. [Online]. Available: \url{https://doi.org/10.1145/3065386}
\BIBentrySTDinterwordspacing

\bibitem{Pascal}
M.~Everingham, L.~Van~Gool, C.~Williams, J.~Winn, and A.~Zisserman, ``The
  pascal visual object classes (voc) challenge,'' \emph{Int. J. Comput. Vis.},
  vol.~88, pp. 303--338, 06 2010.

\bibitem{MSCOCO}
T.-Y. Lin, M.~Maire, S.~J. Belongie, J.~Hays, P.~Perona, D.~Ramanan,
  P.~Doll{\'a}r, and C.~L. Zitnick, ``Microsoft coco: Common objects in
  context,'' in \emph{Proc. Europ. Conf. Comp.}, 2014, p. 740–755.

\bibitem{RCNN}
R.~{Girshick}, J.~{Donahue}, T.~{Darrell}, and J.~{Malik}, ``Region-based
  convolutional networks for accurate object detection and segmentation,''
  \emph{IEEE Trans. Pattern Anal. Mach. Intell.}, vol.~38, no.~1, pp. 142--158,
  2016.

\bibitem{Girshick2015FastR}
R.~B. Girshick, ``Fast r-cnn,'' \emph{Proc. IEEE Int. Conf. Comput. Vis.}, pp.
  1440--1448, 2015.

\bibitem{FasterR}
S.~{Ren}, K.~{He}, R.~{Girshick}, and J.~{Sun}, ``Faster r-cnn: Towards
  real-time object detection with region proposal networks,'' \emph{IEEE Trans.
  Pattern Anal. Mach. Intell.}, vol.~39, no.~6, pp. 1137--1149, 2017.

\bibitem{Liu2016SSDSS}
W.~Liu, D.~Anguelov, D.~Erhan, C.~Szegedy, S.~Reed, C.-Y. Fu, and A.~Berg,
  ``Ssd: Single shot multibox detector,'' in \emph{Proc. Europ. Conf. Comp.},
  2016.

\bibitem{YOLO}
J.~{Redmon}, S.~{Divvala}, R.~{Girshick}, and A.~{Farhadi}, ``You only look
  once: Unified, real-time object detection,'' in \emph{Proc. IEEE Conf.
  Comput. Vis. and Pattern Recog.}, 2016, pp. 779--788.

\bibitem{8125749}
Y.~Li, S.~Tang, M.~Lin, Y.~Zhang, J.~Li, and S.~Yan, ``Implicit negative
  sub-categorization and sink diversion for object detection,'' \emph{IEEE
  Trans. Image Process}, vol.~27, no.~4, pp. 1561--1574, 2018.

\bibitem{9156611}
Y.~Li, T.~Wang, B.~Kang, S.~Tang, C.~Wang, J.~Li, and J.~Feng, ``Overcoming
  classifier imbalance for long-tail object detection with balanced group
  softmax,'' in \emph{Proc. IEEE Conf. Comput. Vis. and Pattern Recog.}, 2020,
  pp. 10\,988--10\,997.

\bibitem{DETR}
N.~Carion, F.~Massa, G.~Synnaeve, N.~Usunier, A.~Kirillov, and S.~Zagoruyko,
  ``End-to-end object detection with transformers,'' \emph{ArXiv}, vol.
  abs/2005.12872, 2020.

\bibitem{WSDDN}
H.~Bilen and A.~Vedaldi, ``Weakly supervised deep detection networks,'' in
  \emph{Proc. IEEE Conf. Comput. Vis. and Pattern Recog.}, Jun 2016, pp.
  2846--2854.

\bibitem{PCL}
P.~Tang, X.~Wang, S.~Bai, W.~Shen, X.~Bai, W.~Liu, and A.~Yuille, ``Pcl:
  Proposal cluster learning for weakly supervised object detection,''
  \emph{IEEE Trans. Pattern Anal. Mach. Intell.}, vol.~42, no.~1, p. 176–191,
  Jan 2020.

\bibitem{OIM}
C.~Lin, S.~Wang, D.~Xu, Y.~Lu, and W.~Zhang, ``Object instance mining for
  weakly supervised object detection,'' in \emph{Proc. AAAI Conf. Artif.
  Intell.}, 2020, p. 11482–11489.

\bibitem{Yang2019TowardsPE}
K.~Yang, D.~Li, and Y.~Dou, ``Towards precise end-to-end weakly supervised
  object detection network,'' in \emph{Proc. IEEE Int. Conf. Comput. Vis.},
  2019, pp. 8371--8380.

\bibitem{WSOD2}
Z.~Zeng, B.~Liu, J.~Fu, H.~Chao, and L.~Zhang, ``Wsod2: Learning bottom-up and
  top-down objectness distillation for weakly-supervised object detection,'' in
  \emph{Proc. IEEE Int. Conf. Comput. Vis.}, October 2019, pp. 8291--8299.

\bibitem{Ren2020InstanceawareCA}
Z.~Ren, Z.~Yu, X.~Yang, M.~Liu, Y.~Lee, A.~G. Schwing, and J.~Kautz,
  ``Instance-aware, context-focused, and memory-efficient weakly supervised
  object detection,'' in \emph{Proc. IEEE Conf. Comput. Vis. and Pattern
  Recog.}, 2020, pp. 10\,595--10\,604.

\bibitem{Gaidon11actomsequence}
A.~Gaidon, Z.~Harchaoui, and C.~Schmid, ``Actom sequence models for efficient
  action detection,'' in \emph{Proc. IEEE Conf. Comput. Vis. and Pattern
  Recog.}, 2011, pp. 3201--3208.

\bibitem{Zhao17}
Y.~Zhao, Y.~Xiong, L.~Wang, Z.~Wu, X.~Tang, and D.~Lin, ``Temporal action
  detection with structured segment networks,'' in \emph{Proc. IEEE Int. Conf.
  Comput. Vis.}, 2017, pp. 2933--2942.

\bibitem{Lin17}
T.~Lin, X.~Zhao, and Z.~Shou, ``Single shot temporal action detection,'' in
  \emph{Proc. ACM Multimedia Conf.}, 2017, p. 988–996.

\bibitem{Chao2018RethinkingTF}
Y.-W. Chao, S.~Vijayanarasimhan, B.~Seybold, D.~Ross, J.~Deng, and
  R.~Sukthankar, ``Rethinking the faster r-cnn architecture for temporal action
  localization,'' 2018, pp. 1130--1139.

\bibitem{SCNN}
\BIBentryALTinterwordspacing
Z.~Shou, D.~Wang, and S.~Chang, ``Temporal action localization in untrimmed
  videos via multi-stage cnns,'' in \emph{Proc. IEEE Conf. Comput. Vis. and
  Pattern Recog.}\hskip 1em plus 0.5em minus 0.4em\relax Los Alamitos, CA, USA:
  IEEE Computer Society, jun 2016, pp. 1049--1058. [Online]. Available:
  \url{https://doi.ieeecomputersociety.org/10.1109/CVPR.2016.119}
\BIBentrySTDinterwordspacing

\bibitem{TAG}
Y.~Xiong, Y.~Zhao, L.~Wang, D.~Lin, and X.~Tang, ``A pursuit of temporal
  accuracy in general activity detection,'' 2017.

\bibitem{TURN-TAP}
J.~Gao, Z.~Yang, K.~Chen, C.~Sun, and R.~Nevatia, ``{TURN TAP}: Temporal unit
  regression network for temporal action proposals,'' in \emph{Proc. IEEE Int.
  Conf. Comput. Vis.}, Oct 2017.

\bibitem{UntrimmedNet}
L.~Wang, Y.~Xiong, D.~Lin, and L.~Gool, ``Untrimmednets for weakly supervised
  action recognition and detection,'' in \emph{Proc. IEEE Conf. Comput. Vis.
  and Pattern Recog.}, 2017, pp. 6402--6411.

\bibitem{Singh2017HideandSeekFA}
K.~K. Singh and Y.~Lee, ``Hide-and-seek: Forcing a network to be meticulous for
  weakly-supervised object and action localization,'' \emph{Proc. IEEE Int.
  Conf. Comput. Vis.}, pp. 3544--3553, 2017.

\bibitem{STPN}
P.~Nguyen, T.~Liu, G.~Prasad, and B.~Han, ``Weakly supervised action
  localization by sparse temporal pooling network,'' in \emph{Proc. IEEE Conf.
  Comput. Vis. and Pattern Recog.}, 2018, pp. 6752--6761.

\bibitem{AutoLoc}
Z.~Shou, H.~Gao, L.~Zhang, K.~Miyazawa, and S.~Chang, ``Autoloc:
  Weakly-supervised temporal action localization in untrimmed videos,'' in
  \emph{Proc. Europ. Conf. Comp.}, 2018.

\bibitem{Nguyen2019WeaklySupervisedAL}
P.~Nguyen, D.~Ramanan, and C.~C. Fowlkes, ``Weakly-supervised action
  localization with background modeling,'' \emph{Proc. IEEE Int. Conf. Comput.
  Vis.}, pp. 5501--5510, 2019.

\bibitem{ROLE2018}
M.~Liu, X.~Wang, L.~Nie, Q.~Tian, B.~Chen, and T.-S. Chua, ``Cross-modal moment
  localization in videos,'' in \emph{Proc. ACM Multimedia Conf.}\hskip 1em plus
  0.5em minus 0.4em\relax Association for Computing Machinery, 2018, p.
  843–851.

\bibitem{ABLR2018}
Y.~Yuan, T.~Mei, and W.~Zhu, ``To find where you talk: Temporal sentence
  localization in video with attention based location regression,'' in
  \emph{Proc. AAAI Conf. Artif. Intell.}, vol.~33, 04 2018.

\bibitem{Xu2019MultilevelLA}
H.~Xu, K.~He, B.~A. Plummer, L.~Sigal, S.~Sclaroff, and K.~Saenko, ``Multilevel
  language and vision integration for text-to-clip retrieval,'' in \emph{Proc.
  AAAI Conf. Artif. Intell.}, 2019, p. 9062–9069.

\bibitem{aaai19Chen}
S.~Chen and Y.-G. Jiang, ``Semantic proposal for activity localization in
  videos via sentence query,'' in \emph{Proc. AAAI Conf. Artif. Intell.}, 2019,
  pp. 8199--8206.

\bibitem{SIGIRZhang}
Z.~Zhang, Z.~Lin, Z.~Zhao, and Z.~Xiao, ``Cross-modal interaction networks for
  query-based moment retrieval in videos,'' in \emph{Proc. Int. ACM SIGIR Conf.
  Res. Dev. Inf. Retr.}, 2019, pp. 3750--3762.

\bibitem{Glove}
J.~Pennington, R.~Socher, and C.~Manning, ``Glove: Global vectors for word
  representation,'' in \emph{Conf. Empir. Methods Nat. Lang. Process., Proc.
  Conf.}, vol.~14, 01 2014, pp. 1532--1543.

\bibitem{LSTM}
S.~Hochreiter and J.~Schmidhuber, ``Long short-term memory,''
  \emph{Ncomputing}, vol.~9, no.~8, p. 1735–1780, Nov. 1997.

\bibitem{CIAN}
J.~Fan, Z.~Zhang, T.~Tan, C.~Song, and J.~Xiao, ``Cian: Cross-image affinity
  net for weakly supervised semantic segmentation,'' in \emph{Proc. AAAI Conf.
  Artif. Intell.}, Apr 2020, p. 10762–10769.

\bibitem{Charades}
G.~A. Sigurdsson, G.~Varol, X.~Wang, A.~Farhadi, I.~Laptev, and A.~Gupta,
  ``Hollywood in homes: Crowdsourcing data collection for activity
  understanding,'' in \emph{Proc. Europ. Conf. Comp.}, 2016, pp. 510--526.

\bibitem{VGG}
K.~Simonyan and A.~Zisserman, ``Very deep convolutional networks for
  large-scale image recognition,'' 2014.

\bibitem{C3D}
D.~{Tran}, L.~{Bourdev}, R.~{Fergus}, L.~{Torresani}, and M.~{Paluri},
  ``Learning spatiotemporal features with 3d convolutional networks,'' in
  \emph{Proc. IEEE Int. Conf. Comput. Vis.}, 2015, pp. 4489--4497.

\bibitem{TGN}
J.~Chen, X.~Chen, L.~Ma, Z.~Jie, and T.-S. Chua, ``Temporally grounding natural
  sentence in video,'' in \emph{Proc. Conf. Empir. Methods Nat. Lang. Process.,
  EMNLP}, Oct.-Nov. 2018, pp. 162--171.

\end{thebibliography}
\end{document}